\documentclass[runningheads]{llncs}
\usepackage{amsmath}
\usepackage{amssymb}
\usepackage{graphicx}
\usepackage{subcaption}
\usepackage{booktabs}
\usepackage{xcolor}

\begin{document}
\title{RATCHET: Medical Transformer for \\ Chest X-ray Diagnosis and Reporting}
\titlerunning{RATCHET: Medical Transformer for Chest X-ray Diagnosis and Reporting}

\author{
  Benjamin Hou$^{1}$, 
  Georgios Kaissis$^{2}$, 
  Ronald M. Summers$^{4}$, 
  Bernhard Kainz$^{1,3}$
}


\authorrunning{B. Hou et al.}

\institute{
  $^{1}$Imperial College London, UK, 
  $^{2}$Technische Universität München, DE, \\ 
  $^{3}$FAU Erlangen--N\"urnberg, DE,
  $^{4}$National Institutes of Health, USA 
}

\maketitle              
\begin{abstract}
Chest radiographs are one of the most common diagnostic modalities in clinical routine. It can be done cheaply, requires minimal equipment, and the image can be diagnosed by every radiologists. However, the number of chest radiographs obtained on a daily basis can easily overwhelm the available clinical capacities.
We propose RATCHET: RAdiological Text Captioning for Human Examined Thoraces. RATCHET is a CNN-RNN-based medical transformer that is trained end-to-end. It is capable of extracting image features from chest radiographs, and generates medically accurate text reports that fit seamlessly into clinical work flows. The model is evaluated for its natural language generation ability using common metrics from NLP literature, as well as its medically accuracy through a surrogate report classification task. The model is available for download at: \url{http://www.github.com/farrell236/RATCHET}. 

\end{abstract}

\section{Introduction}

Automatic report generation for clinical chest radiographs have recently attracted attention due to advances in Natural Language Processing (NLP) and the introduction of transformers. In the past decade, learning-based models for radiographic applications have evolved from single image classification~\cite{DBLP:conf/cbms/XueYCJALT15,DBLP:conf/crv/KumarWC15}, multi-label 
classification~\cite{DBLP:journals/corr/abs-1711-05225,DBLP:conf/midl/ChenMXHH19}, to more complex systems that incorporates a mixture of modalities~\cite{DBLP:journals/corr/abs-2010-01165}, uncertainty~\cite{DBLP:conf/cvpr/YangYPCSGY19} and explainability~\cite{pasa2019efficient}.


Learning-based methods often require large and well annotated datasets, something that is available in computer vision, e.g. MS-COCO~\cite{DBLP:conf/eccv/LinMBHPRDZ14}, but not in medical imaging. Generic large-scale annotated medical datasets are more difficult to acquire, as; (i) clinical experts need to devote extra time to annotate images,  (ii) the images are likely to be bound by specific scanning parameters from the imaging device, and (iii) clinicians commonly annotate via free-text instead of single class labels during routine diagnostics. 
Medical reports are often preferred for diagnostics, as prose can describe a pathology in more detail compared to a single image label. 
However, clinical report writing is tedious, highly variable, and a time consuming task. In countries where the population is large and radiologists are in high demand, a single radiologist may be required to read hundreds of radiology images and/or reports per day~\cite{DBLP:journals/artmed/MonshiPC20}.

We seek to develop a model that can perform automatic report generation for chest radiographs with high \emph{medical} accuracy. Our intention is not to replace radiologists, but instead to assist them in their day-to-day clinical duties by accelerating the process of report generation. Ultimately, the attending physician will be required to usurp responsibility of the final diagnosis, as bounded by ongoing debate in the ethics of AI, but this is beyond the scope of this paper.


\noindent\textbf{Contribution:}
Inspired by Neural Machine Translation, we develop a transformer based CNN-Encoder to RNN-Decoder architecture for generating chest radiograph reports. We use attention to localize regions of interest in the image and show where the network is focusing on for the corresponding generated text. This is to reinforce explainability in black-box models for clinical settings, as well as a means for extracting bounding boxes for disease localization. We evaluate the model for both, its natural language capability, and more importantly, the medical accuracy of generated reports.

\noindent\textbf{Related Works:}
The first transformer-based network for Neural Machine Translation (NMT) was introduced by Vaswani et al~\cite{DBLP:conf/nips/VaswaniSPUJGKP17}. Here, the encoder and decoder LSTM networks are replaced by stacks of Masked Multi-Head Attention with Point-Wise Feed Forward Network modules that operate on whole sequences directly. The architecture decouples the time dependency of RNNs, which allows for simultaneous processing of sequential data and alleviates the problem of vanishing gradients (a well known problem with RNNs~\cite{DBLP:conf/icml/PascanuMB13}).
Transformer networks can be trained much faster, and able to achieve much higher accuracy with fewer data as the attention mechanism learns contextual relations between words and sub-words all in one sequence.


Numerous works have explored radiology report generation~\cite{DBLP:journals/artmed/MonshiPC20}. Wang et al~\cite{DBLP:conf/cvpr/WangPLLS18} introduced TieNet, a CNN-RNN architecture that combines image saliency maps with attention-encoded text embedding to jointly learn the disease class and report generation. 
Liu et al~\cite{DBLP:conf/mlhc/LiuHMBWSG19} later proposed a Hierarchical Generation method using a CNN-RNN-RNN architecture, with a Reinforcement Learning reward signal for better readability and language coherency. At the same time, Boag et al~\cite{DBLP:conf/nips/BoagHMBAS19} performed several baseline experiments for chest radiograph report generation. This explored and compared three main report generation techniques: Conditional n-gram Language Model, Nearest Neighbor (similar to data retrieval, where the model recalls the caption with the largest cosine similarity to the query image), and `Show-and-Tell'~\cite{DBLP:conf/cvpr/VinyalsTBE15}.
Yuan et al~\cite{DBLP:conf/miccai/YuanLLL19} proposed a hierarchical approach with two nested LSTM decoders. The first decoder generates sentence hidden states from image features, whereas the second decoder generates individual words from the sentence hidden states. The image encoder also leverages additional information during the training process in the form of frontal and lateral image input, as well as image labels and Medical Text Indexer (MTI) Tags.
Li et al~\cite{DBLP:conf/aaai/LiLHX19} also used a hierarchical approach for report generation, but opted for four Graph Transformer (GTR) Networks instead of CNNs and LSTMs. The input image features get encoded by a GTR to an abnormality graph, which then gets transformed by another GTR to a disease graph that yields labels specific to the input. In parallel, two more GTRs are used to convert the abnormality graph to a report template, and then subsequently to a text report.
Syeda-Mahmood et al~\cite{DBLP:conf/miccai/Syeda-MahmoodWG20} then simplified the process by training a custom deep learning network on Fine Finding Labels (FFL), which consists of four elements: the finding type, positive or negative finding, core finding, and modifiers. This constrains the network to focus only on the key and relevant pathological details. 




\section{Method}

The process of report generation starts with tokenization of the text, which allows the machine to grasp the concept of text as numbers. There are three key aspects to training the report generation model: (i) text pre-processing, (ii) tokenization and (iii) language model formulation/training. 
A report consists of three parts: The \emph{Findings} section details what the radiologist saw in each area of the body during the exam, whereas the \emph{Impressions} is a summary of all Findings with the patients clinical history, symptoms, and reason for the exam taken into account. It is possible for the Findings section to be empty, which means that the radiologist did not find any problems to report back to the doctor in charge. 

\begin{figure}
\centering
\includegraphics[width=0.8\textwidth]{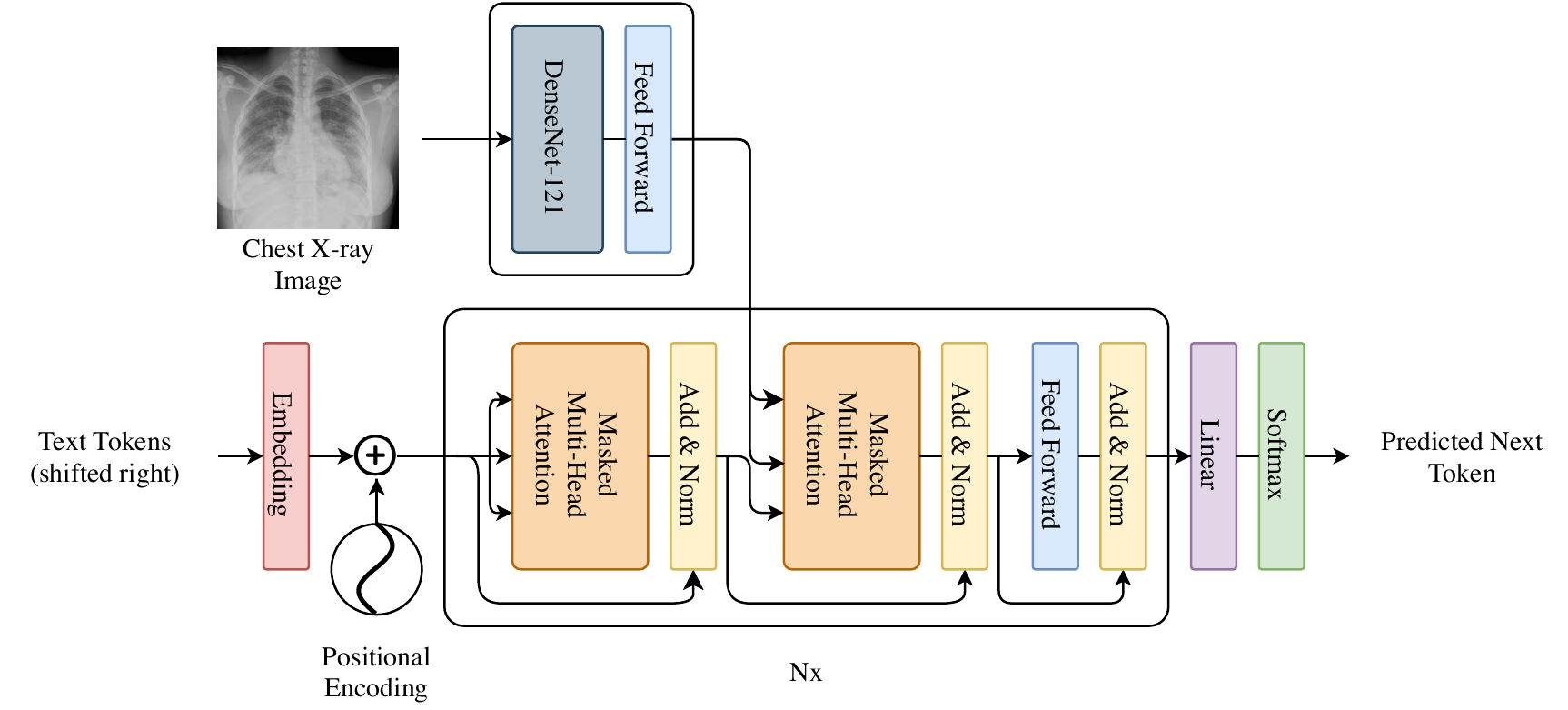}
\caption{Model Architecture for RATCHET.}
\label{fig:exp_pipeline}
\end{figure}

\noindent\textbf{Language Pre-processing and Tokenization:}
Jargon heavily impact the tokenization process, and subsequently the ability of the model to learn. Na\"ive tokenizers require the input sentence to be vigorously cleaned, \emph{e.g.}, splitting of whole words, punctuation, measurements, units, abbreviations, etc.. This can be especially problematic in medical texts as medical etymology involves compounding multiple Latin affixes, thus creating many unique words, as well as measurements. Too many infrequent tokens can hamper the model's ability to learn, whereas pruning too many infrequent tokens using the Out-Of-Vocabulary (OOV) token results in a stunted vocabulary. 


Alternatively, a more complex tokenizer can be used, e.g., BPE~\cite{DBLP:conf/acl/SennrichHB16a},
where words are broken down into one or more sub-words. This removes the dependency on the OOV token, and can also incorporate punctuation into the vocabulary corpus. Both sub-word tokenization methods share the same fundamental idea:  more frequent words should be given unique IDs, whereas less frequent words should be decomposed into sub-words that best retain their meaning. In this work, the Huggingface \includegraphics[height=0.8em]{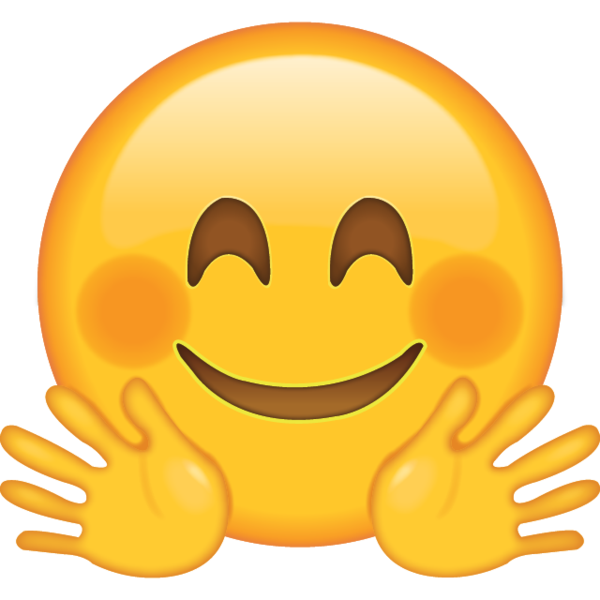} library~\cite{DBLP:journals/corr/abs-1910-03771} is used to create custom BPE tokenizers. 



\noindent\textbf{Captioning Transformer:}
RATCHET, as shown in Figure~\ref{fig:exp_pipeline}, follows the Encoder - Decoder architecture style for NMT. Encouraged by results in CheXNet~\cite{DBLP:journals/corr/abs-1711-05225}, the encoder module is replaced with a DenseNet-121~\cite{DBLP:conf/cvpr/HuangLMW17} as the primary image feature extractor. The decoder architecture remains the same as a text transformer, and is used in an auto-regressive manner. The output feature shape of DenseNet-121, for an input image of size $224\times224$, is $7\times7\times1024$. This is flattened to a vector of size $49\times1024$, which allows it to be treated the same way as a sequence of text. The mask for the self-attention sub-layer in the decoder stack is disabled for the encoder input, this ensures that the predictions for feature pixel $i$ can depend on any or all other feature pixels. 

\noindent\textbf{Scaled Dot-Product Attention:}
\cite{DBLP:conf/nips/VaswaniSPUJGKP17} is a method of allowing the network to focus on a localized set of values from the input. Inspired by retrieval systems, the module features three inputs; $\boldsymbol{Q}$uery, $\boldsymbol{K}$ey, $\boldsymbol{V}$alue.
\begin{equation}
    Attention(Q,K,V) = softmax(\frac{QK^T}{\sqrt{d_k}})V,  
\end{equation}
where $Q \in \mathbb{R}^{L \times d}$, $K \in \mathbb{R}^{L \times d}$, $V \in \mathbb{R}^{L \times d_v}$, $L$ is the sequence length and $d$ is the features depth. The module takes the query, finds the most similar key and returns the value that correspond to this key. This is accelerated by two matrix multiplications and a softmax operation.  $softmax(QK^T)$ creates a probability distribution that peaks at places localized by the keys for the corresponding query. This acts as a pseudo mask, and by matrix multiplying it with $V$ returns the localized values the network should focus on. 

\noindent\textbf{Masked Multi-Head Attention:}
The M-MHA module is composed of $n$-stacked Scaled Dot-Product Attention modules in parallel, and is defined by:
\begin{align*}
    h_i &= Attention(QW_i^Q, KW_i^K, VW_i^V), \\
    H &= Concat(h_1, h_2, ..., h_n), \\
    O &= HW_h,
\end{align*}
where $W_i^Q \in \mathbb{R}^{d_{model} \times d}$, $W_i^K \in \mathbb{R}^{d_{model} \times d}$, and $W_i^V \in \mathbb{R}^{d_{model} \times d_v}$. The attention output of each head, $h_i$, is concatenated together and projected to the same cardinality as the input by multiplying it with $W_h$, such that $W_h \in \mathbb{R}^{(n \times d_v) \times d_{model}}$ and $O \in \mathbb{R}^{L \times d_{model}}$.

The first M-MHA in the decoder is set in self-attention mode, \emph{i.e.}, $K=Q=V=\text{input}$. $K$ and $V$ behave like a memory component of the network. It recalls the appropriate region to focus on for a given a query input based on words and sequences seen during training.
The second M-MHA takes the image features as $K$ and $Q$ inputs, with the text features as $V$. The attention module then picks the necessary text features based on the image features, and simultaneously merges the image modality with the text modality. This module is used for model explainability, where highlighted features indicates the region of the image the model is particularly focusing on w.r.t. the text generated.


\section{Experiment and Results}

\noindent\textbf{Dataset:} MIMIC CXR v2.0.0~\cite{DBLP:journals/corr/abs-1901-07042} is a recently released dataset of chest radiographs, with free-text radiology reports. It is publicly available and large, containing more than 300,000 images from 60,000 patients. It has been acquired at the Beth Israel Deaconess Medical Center in Boston, MA. The dataset also comes pre-annotated with disease labels, mined using either CheXpert~\cite{DBLP:conf/aaai/IrvinRKYCCMHBSS19} or NegBio~\cite{DBLP:journals/corr/abs-1712-05898} NLP tool, as well as train/validate/test splits. Since the radiology reports are free-text, they are very much prone to typos and inconsistencies that can hamper the efficacy of deep learning algorithms. The dataset has been further processed, with typos and errors corrected~\cite{mimic-cxr-fork}. Only AP/PA views are used for our experiments.
The gold standard ground truth for the dataset is the radiology reports. Due to the extreme time cost of labeling images by hand, the disease class labels are all mined via automatic means.

To ensure a fair testing of architectures, the input and evaluation part of the pipeline remains identical for all experiments. This includes the dataset used, image pre-processing and data augmentation methods, vocabulary construction, tokenization and NLP metrics for evaluation. We re-implemented the original LSTM-based method as presented in~\cite{DBLP:conf/icml/XuBKCCSZB15} for the baseline architecture, and further modified it to create TieNet~\cite{DBLP:conf/cvpr/WangPLLS18}. Since the implementation for TieNet has not been released, we have re-implemented it according to the descriptions provided by the original authors to the best of our abilities. 

\noindent\textbf{Image Input}: As the images in the datasets are all non standard, each image is resized with padding to 224x224. Data augmentation include 
random shifting of brightness, saturation and contrast to mimic various exposure parameters before being rescaled to an intensity range between $[0,1]$ for network training.

\noindent\textbf{Network Training:} All networks are trained using Tensorflow 2.4.1 and Keras on a Nvidia Titan X, using the Adam optimiser with a learning rate of $1 \times 10^{-4}$ and a batch size of 16 for 20 epochs. All weights start from random initialization, with a total parameter count of approx 51M. The hyper-parameters used for all models were; \emph{num\_layers}=6, \emph{n\_head}=8, \emph{d\_model}=512, \emph{dff}=2048, \emph{dropout}=0.2. During training, an image $I$ is fed into the encoder, with the output feature maps flattened into a linear sequence. The decoder is trained in a ``teacher forcing'' technique where the entire ground truth sequence $S \in \mathbb{R}^{L \times n_{vocab}}$ is fed to the decoder to predict $S' \in \mathbb{R}^{L \times n_{vocab}}$. $S'$ is the same as $S$ except with the tokens shifted to the right by one, thus training the model to predict the next likely token of that sequence. 
With text represented by a one-hot vector, i.e., size of the sequence length by number of tokens in vocabulary ($L \times n_{vocab}$), the loss is therefore defined by: 
\begin{equation}
    \log p(S | I) = \sum_{i=0}^{N} \log p(S_t | I , S_0, S_1, ... S_{t-1}, \theta), 
\label{eqn:loss}
\end{equation}
where $(I,S)$ is the Image-Text pair, and $\theta$ are the network parameters. The network is trained by minimizing the cross-entropy loss to optimize for the sum of log probabilities across all $N$ training examples. 
Inference takes approx. 0.12s.

\begin{figure}[h]
\centering
\includegraphics[width=0.8\textwidth]{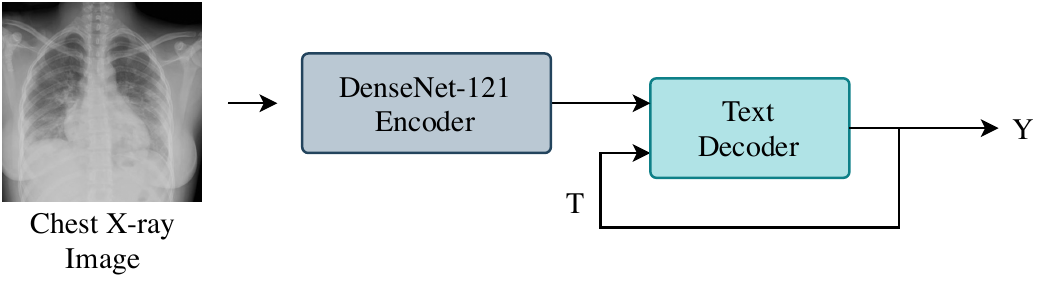}
\caption{Running Inference using RATCHET}
\label{fig:inference}
\end{figure}

\noindent\textbf{Model Inference:} To generate text, RATCHET is ran in an auto-regressive fashion. For a given chest xray, it is resized to $224\times224$ with padding, and intensity rescaled to range of $[0,1]$. The image is then fed into the model along with the initial token, \verb|<BOS>| (Beginning Of Sentence). RATCHET then predicts the next token in accordance to the input it's received. The predicted token is then concatenated with the previous token to form a sequence, and the model is ran again, as shown in Figure~\ref{fig:inference}. This is repeated until either a \verb|<EOS>| (End Of Sentence) token is predicted, or it has reached a maximum sequence length of 128 tokens, as bounded by the longest radiology report in training dataset. 

\noindent\textbf{Results:}
Table~\ref{tab:consresults} shows a summary of the average NLP metric scores for all predicted text compared to the ground truth radiology reports. A more detailed table containing all classes can be found in the supplementary materials. We attain better report language quality scores than both baseline method and TieNet. Interestingly, TieNet scored just sub-par compared to the baseline method. This could be attributed to the fact that TieNet was originally trained on the OpenI~\cite{DBLP:journals/jamia/Demner-FushmanK16} dataset, which had 5x fewer vocabulary tokens compared to MIMIC-CXR, and non-optimal hyper-parameters as it had additional classification loss.

\begin{table}[t]
\caption{NLP evaluation results on MIMIC-CXR Report Generation. A  table with  more detailed results can be found in the supplemental material. (B-1: BLEU-1, MET.:  METEOR, R\_L: ROUGE\_L scores, Cm: Cardiomegaly, Ed: Edema, Co: Consolidation, At: Atelectasis, PE: Pleural Effusion. The latter are selected classes overlapping with CheXNet.) Note that the classification task has been evaluated as a surrogate to validate medical accuracy of the generated reports. We would not expect report classification to outperform direct prediction from images. Best language quality performance in bold.}
\label{tab:consresults}
\centering
\scalebox{0.95}{
\begin{tabular}{lcccccccccc}
\toprule
Method & \multicolumn{5}{c}{NLP Metrics} & \multicolumn{5}{c}{Classification F1} \\
\cmidrule(lr){2-6} \cmidrule(lr){7-11}
 & B-1 & MET. & R\_L & CIDEr & SPICE & Cm & Ed & Co & At & PE \\
 \cmidrule(lr){2-6} \cmidrule(lr){7-11}
CheXNet (images) &-&-&-&-&-&0.534&0.674&0.193&0.567&0.715\\
\midrule
TieNet (Classification) &-&-&-&-&-&0.488&0.536&0.188&0.471&0.591\\
Baseline (NLP) &0.208&\textbf{0.108}&0.217&0.419&0.107&-&-&-&-&-\\
TieNet (NLP) &0.190&0.069&0.200&0.411&0.084&0.061&0.050&0.0&0.006&0.033\\
RATCHET (NLP) &\textbf{0.232}&0.101&\textbf{0.240}&\textbf{0.493}&\textbf{0.127}&0.446&0.407&0.041&0.411&0.633\\
\midrule
Class Bias &-&-&-&-&-&0.221&0.177&0.066&0.245&0.261\\
\bottomrule
\end{tabular}%
}
\end{table}

To assess the medical accuracy of generated reports quantitatively, discrete pathology labels of the ground truth text and generated text were obtained using Stanford's CheXpert Labeler~\cite{irvin2019chexpert}. 
To ensure methodological effectiveness in radiological report generation, this was also compared against a separate direct image classifier, which was trained to perform direct multi-label classification. Following CheXNet~\cite{DBLP:journals/corr/abs-1711-05225}, a DenseNet-121 was trained using the Adam optimizer with a learning rate of $1\times10^{-4}$ and a batch size of 16 for 10 iterations. The epoch with the best performing weights on the validation set was used. Table~\ref{tab:consresults} shows the accuracy of generated reports by the Baseline, TieNet, RATCHET and CheXNet. It can be seen that the results are on par with each other, and with related works in literature~\cite{DBLP:conf/nips/BoagHMBAS19,DBLP:conf/mlhc/LiuHMBWSG19}.

Of the 14 classes, CheXpert only presents the results from radiologists and original labels on ROC curves for 5 classes. Under the defined experiment setting `LabelU' (where uncertain labels are classified as positive), the dataset is still heavily imbalanced at a ratio of 1:4 positive to negative. This is evidently seen in `Consolidation' as the worst performing class across all models. For TieNet, the radiograph report generation head has failed to learn as shown by poor F1 scores. However, the classification head performs just under-par of CheXNet. TieNet's classification head was able to outperform 4 of 5 classes compared to the baseline model, only equaling in performance with `Cardiomegaly'. RATCHET was able to outperform both baseline and TieNet, and also equaling in performance in `Cardiomegaly'. 

\begin{figure}[t]
\centering
\includegraphics[width=1.0\textwidth]{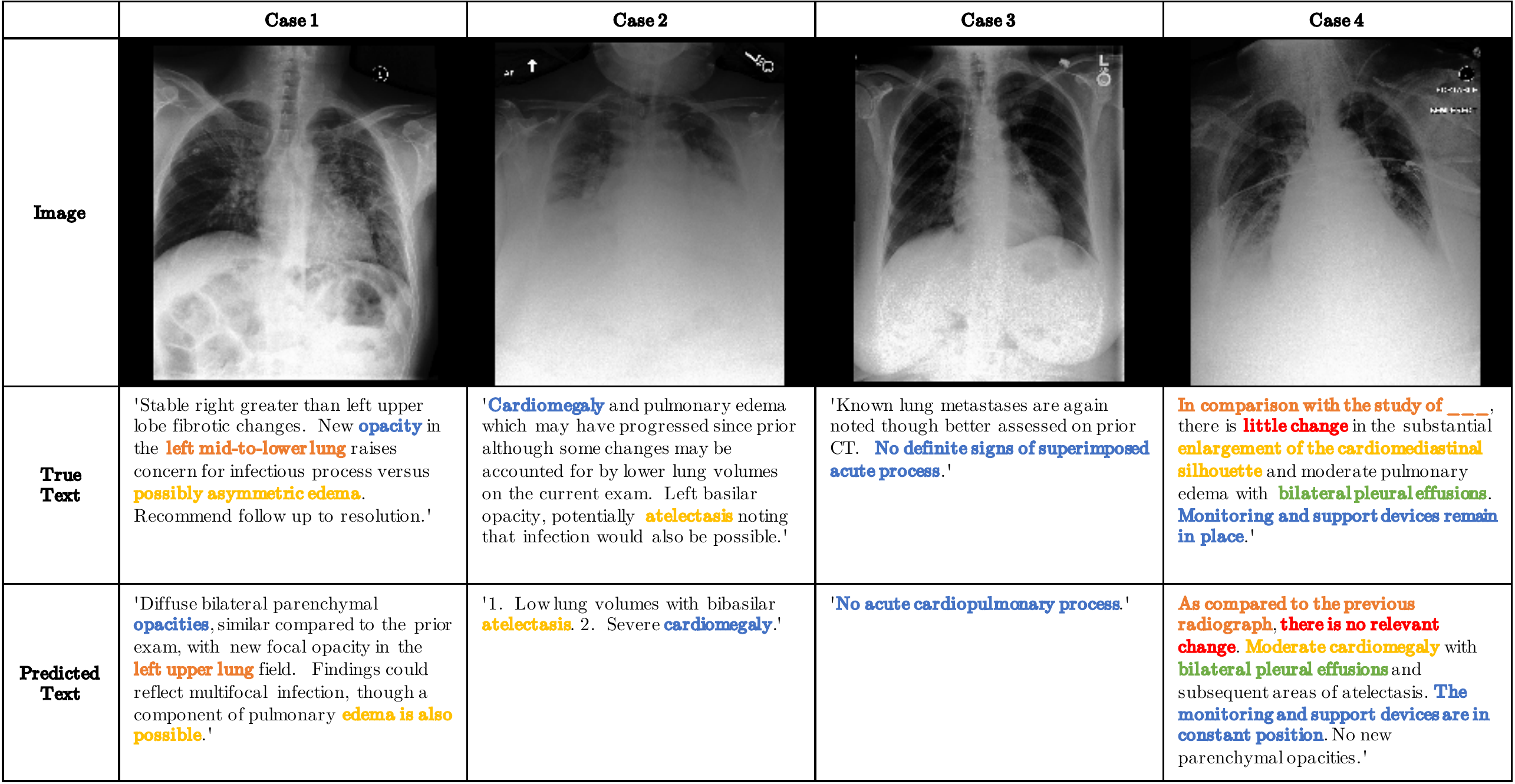}
\caption{Generated Reports for Four Example Test Subjects by RATCHET}
\label{fig:test_set_examples}
\end{figure}

Figure~\ref{fig:test_set_examples} shows qualitative text example reports generated by RATCHET. In each case, it has picked out the correct pathological attributes, despite having very different phrasing. We observe that, at times, all models may preface generated reports with `In comparison with the previous...'. This is a characteristic example of the network trying to imitate~\cite{foster_2007} the training set, a property that was criticized in the GPT model~\cite{marcus_2020}. The network may also need to be better constrained, \emph{e.g.} in Case1, the true text noted `left mid-to-lower lung' whist in the generated text said `left upper lung'. Whilst both the diagnosis of `left' is correct, the network predicted `lower' rather than the true text of `upper'. 
Metrics from the NLP may not be the best for evaluating medical texts, due to specialized tolerances. E.g. in Case4, the true text states `enlargement of the cardiomediastinal silhouette', which the network simplified to `Moderate cardiomegaly'. Whilst this is medically correct, it will yield a lower NLP score.

\begin{figure}[t]
\centering
\captionsetup[subfigure]{labelformat=empty}
\scalebox{0.45}{%
\begin{minipage}{5cm} %
  \centering
  \subfloat[Input test image with cardiomegaly]{\includegraphics[width=5cm]{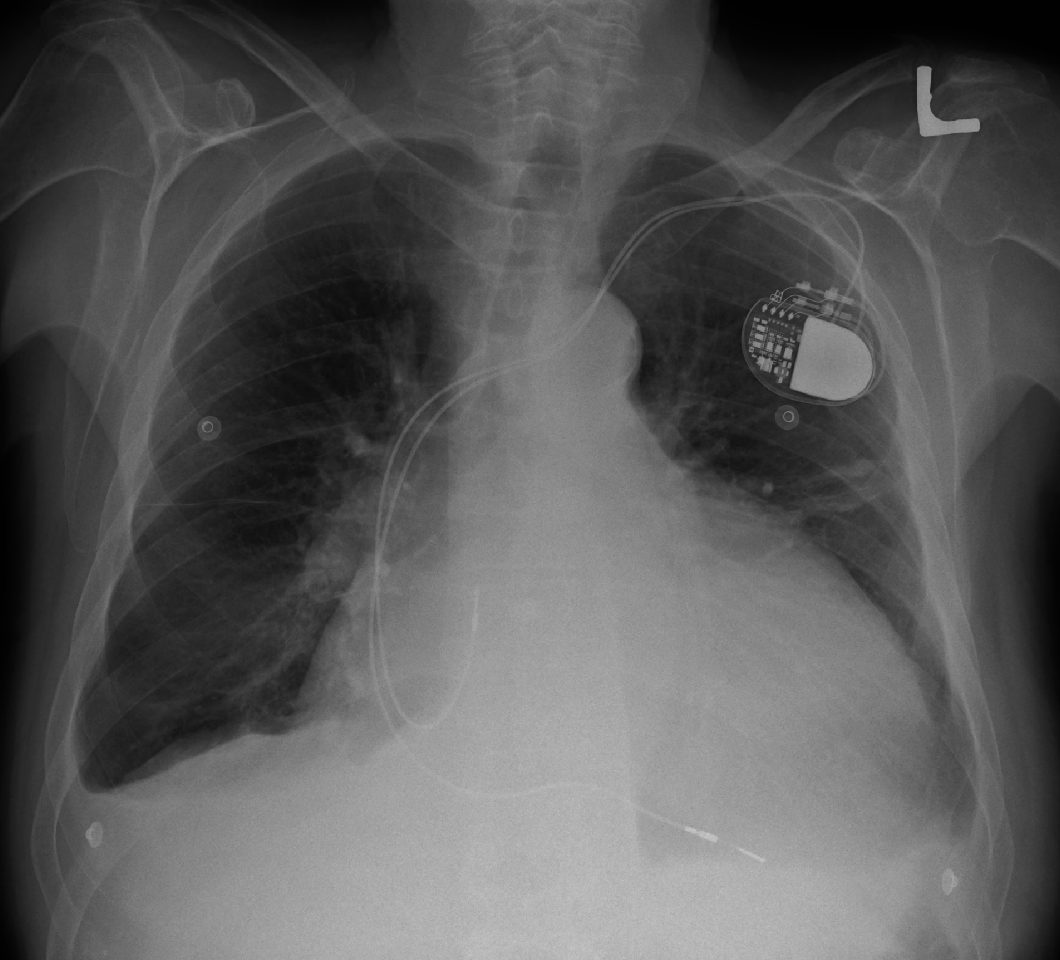}}
\end{minipage} %
\hspace{5mm}
\begin{minipage}{20cm} %
  \centering
  \subfloat[substantial ]{\includegraphics[width=2cm]{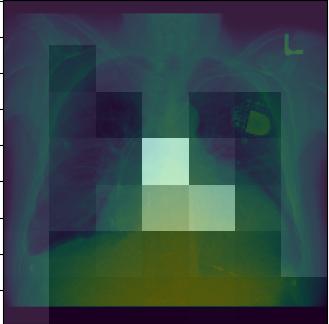}} \hspace{1mm}
  \subfloat[enlargement ]{\includegraphics[width=2cm]{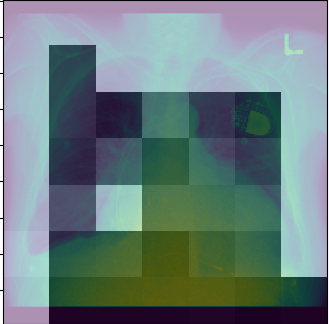}} \hspace{1mm}
  \subfloat[of          ]{\includegraphics[width=2cm]{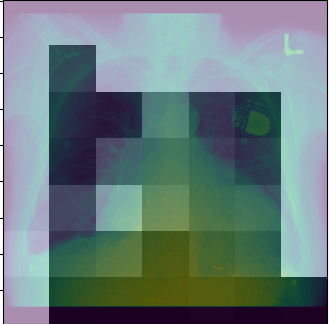}} \hspace{1mm}
  \subfloat[the         ]{\includegraphics[width=2cm]{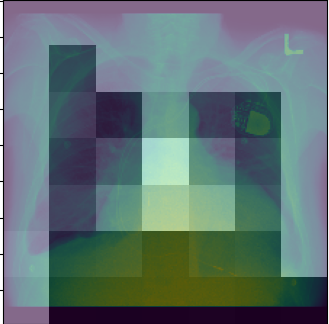}} \hspace{1mm} 
  \subfloat[cardiac     ]{\includegraphics[width=2cm]{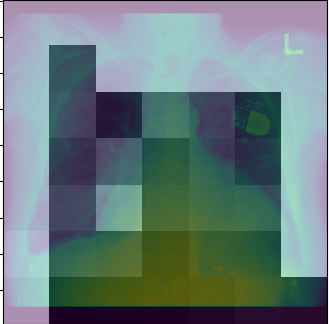}} \hspace{1mm} 
  \subfloat[silhouette  ]{\includegraphics[width=2cm]{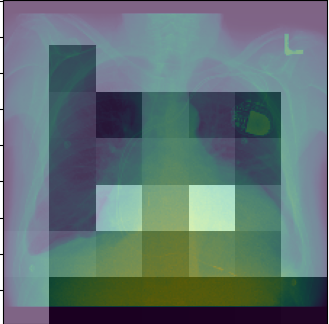}} \hspace{1mm} , \hspace{1mm}
  \subfloat[a           ]{\includegraphics[width=2cm]{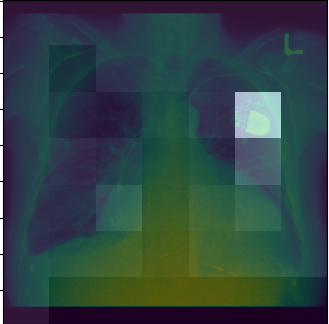}} \hspace{1mm}
  \subfloat[dual        ]{\includegraphics[width=2cm]{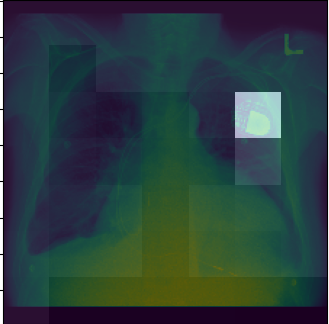}} \hspace{1mm}
  \subfloat[-           ]{\includegraphics[width=2cm]{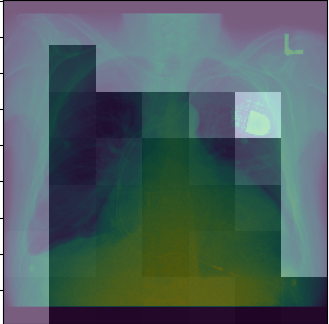}} \hspace{1mm} \\
  \vspace{2mm}
  \subfloat[channel     ]{\includegraphics[width=2cm]{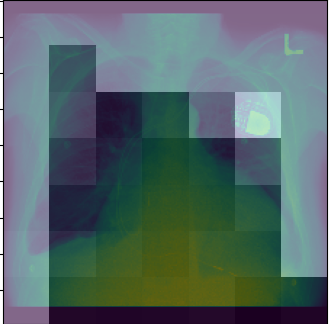}} \hspace{1mm} 
  \subfloat[pacer       ]{\includegraphics[width=2cm]{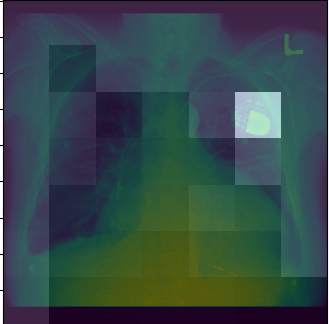}} \hspace{1mm} 
  \subfloat[device      ]{\includegraphics[width=2cm]{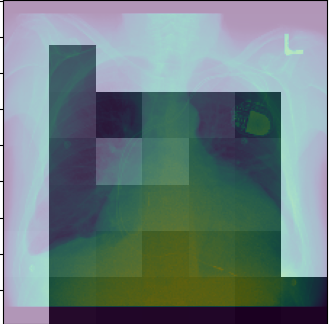}} \hspace{1mm} , \hspace{1mm} 
  \subfloat[Blunting    ]{\includegraphics[width=2cm]{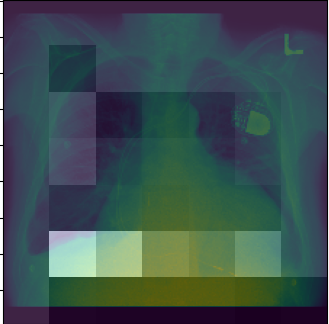}} \hspace{1mm}
  \subfloat[of          ]{\includegraphics[width=2cm]{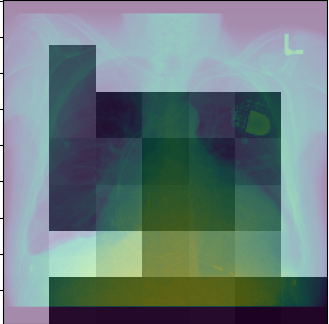}} \hspace{1mm}
  \subfloat[the         ]{\includegraphics[width=2cm]{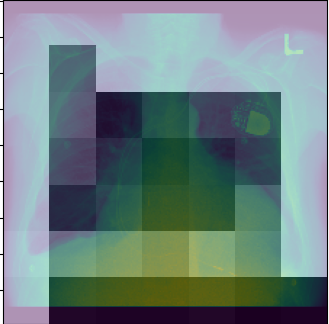}} \hspace{1mm}
  \subfloat[costophrenic]{\includegraphics[width=2cm]{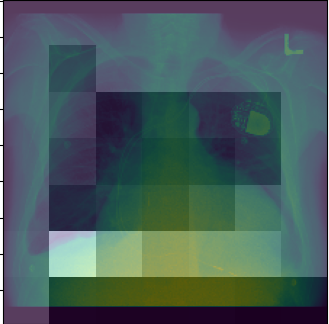}} \hspace{1mm} 
  \subfloat[angles      ]{\includegraphics[width=2cm]{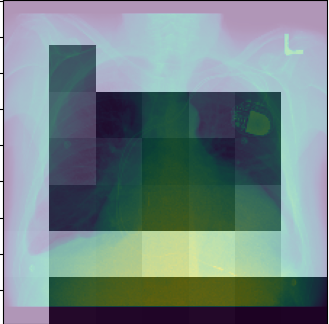}} \hspace{1mm} 
\end{minipage}
}
\caption{A test image with generated caption and attention maps. The generated report for this case contains: \textit{``In comparison with the study of \_\_\_, there is little overall change. Again there is substantial enlargement of the cardiac silhouette with a dual-channel pacer device in place. No evidence of vascular congestion or acute focal pneumonia. Blunting of the costophrenic angles is again seen.''}}

\label{fig:attention_results}
\end{figure}

For model explainability, the attention maps from the second Masked Multi-Head Attention module are used. 
The attention weights are a byproduct of the Scaled Dot-Product Attention, \emph{i.e.} the resultant score of the softmax product between the key and query input matrices. From the entire decoder, only the attention weights from the last stacked decoder module are used, where a reduce mean operation is performed across all Attention Heads. This produces an attention map for each predicted token, Figure~\ref{fig:attention_results} shows the attention of one particular test example. 
It can be seen that the attention module in RATCHET does indeed attend to the appropriate region of the image when referring to certain pieces of text. Notably,  when describing the `enlargement of the heart', the model focuses on the cardiac region. Similarly when `a dual-channel pacer device' is referred to there is a peak in the region where the pacemaker device is located. `Blunting of the costophrenic angles' is the classic sign for pleural effusion, which is a buildup of extra fluid in the space between the lungs and the chest wall (pleural space) at the base of the lungs.

\section{Discussion}
In our experiments, we have consistently shown that a transformer-based architecture can outperform traditional LSTM-based architectures in both NLP linguistic scores as well as medical pathology diagnosis. The attention maps from the Scaled Dot-Product Attention module highlights the region in the image that is responsible for each of the generated text tokens.   

RATCHET, along with other NMT models, are trained on an individual image basis. From a medical point of view, it would be desirable to have it operating on a case by case basis, \emph{i.e.}, to incorporate multiple images, as well as other patient data, clinical patient history and clinical biomarkers potentially including omics information. This would give a more complete picture of the diagnosis as well as constrain the network to make more accurate predictions.


\section{Conclusion}

Text generation is a staple aspect of NLP and remains a challenging task, especially in the realm of medical imaging. RATCHET is a transformer-based CNN-RNN founded on Neural Machine Translation for generating radiological reports. The model is prefaced with a DenseNet-121 model to perform image feature extraction, and is trained end-to-end. In this paper, we have shown that RATCHET is able to generate long medical reports. 

\section{Acknowledgements}

This work is supported by the UK Research and Innovation London Medical Imaging and Artificial Intelligence Centre for Value Based Healthcare, and in part by the Intramural Research Program of the National Institutes of Health Clinical Center.

\bibliographystyle{splncs04}
\bibliography{references}

\end{document}